\documentclass{article}

\PassOptionsToPackage{numbers, compress}{natbib}

\usepackage[preprint]{nips_2018}

\usepackage[utf8]{inputenc} 
\usepackage[T1]{fontenc}    
\usepackage{hyperref}       
\usepackage{url}            
\usepackage{booktabs}       
\usepackage{amsfonts}       
\usepackage{nicefrac}       
\usepackage{microtype}      
\usepackage{tcolorbox}
\usepackage{subfigure}
\usepackage{makecell}

\usepackage{bbm}
\usepackage{bm}
\usepackage{ifthen}
\usepackage{rays_defs_18}
\usepackage{xcolor}

\newcommand\img{\vx}
\newcommand\obs{\vy}
\newcommand\noise{\vw}

\usepackage{xcolor}

\title{Unsupervised Learning with Stein's Unbiased Risk Estimator}

\author{
  Christopher A. Metzler\\
  Stanford University\\
  \texttt{cmetzler@stanford.edu} \\
  \And
  Ali Mousavi\\
  Google AI\\
  \texttt{alimous@google.com} \\
  \And
  Reinhard Heckel\\
  Rice University\\
  \texttt{rh43@rice.edu} \\
  \And
  Richard G. Baraniuk\\
  Rice University\\
  \texttt{richb@rice.edu} \\
}

\begin{document}

\maketitle

\begin{abstract}
Learning from unlabeled and noisy data is one of the grand challenges of machine learning. 
As such, it has seen a flurry of research with new ideas proposed continuously. 
In this work, we revisit a classical idea: Stein's Unbiased Risk Estimator (SURE). 
We show that, in the context of image recovery, SURE and its generalizations can be used to train convolutional neural networks (CNNs) for a range of image denoising and recovery problems {\em without any ground truth data.}

Specifically, our goal is to  reconstruct an image $\img$ from a \emph{noisy} linear transformation (measurement) of the image.
We consider two scenarios: one where no additional data is available and one where we have noisy measurements of other images that belong to the same distribution as $\img$, but have no access to the clean images. 
Such is the case, for instance, in the context of medical imaging, microscopy, and astronomy, where noise-less ground truth data is rarely available.

We show that in this situation, SURE can be used to estimate the mean-squared-error loss associated with an estimate of $\img$. 
Using this estimate of the loss, we train networks to perform denoising and compressed sensing recovery. 
In addition, we also use the SURE framework to partially explain and improve upon an intriguing results presented by Ulyanov et al. in \cite{DeepImagePrior}: that a network initialized with random weights and fit to a single noisy image can effectively denoise that image.

Public implementations of the networks and methods described in this paper can be found at \url{https://github.com/ricedsp/D-AMP_Toolbox}.

\end{abstract}

\section{Introduction}
In this work we consider reconstructing an unknown image $\img \in\mathbb{R}^n$ from measurements of the form $\obs = \mH \img + \noise$, where $\obs\in\mathbb{R}^m$ are the measurements, $\mH\in\mathbb{R}^{m\times n}$ is the linear measurement operator, and $\noise \in\mathbb{R}^m$ denotes noise.
%
This problem arises in numerous applications, including denoising, inpainting, superresolution, deblurring, and compressive sensing.
The goal of an image recovery algorithm is to use prior information about the image's distribution and knowledge of the measurement operator $\mH$ to reconstruct $\img$.

The key determinant of an image recovery algorithm's accuracy is the accuracy of its prior. 
Accordingly, over the past decades a large variety of increasingly complex image priors have been considered, ranging from simple sparse models \cite{SureShrink}, to non-local self-similarity priors \cite{BM3D}, all the way to neural network based priors, in particular CNN based priors \cite{DnCNN}. Among these methods, CNN priors often offer the best performance.
It is widely believed that key to the success of CNN based priors is the ability to process and learn from vast amounts of training data, although recent work suggests the structure of a CNN itself encodes strong prior information \cite{DeepImagePrior}.

CNN image recovery methods are typically trained by taking a representative set of images $\img_1, \img_2, ... \img_L$, drawn from the same distribution as $\img$, and capturing a set of measurements $\obs_1,\obs_2 ... \obs_L$, either physically or in simulation.
The network then learns the mapping $f_\theta: \obs\rightarrow \hat{\img}$ from observations back to images by minimizing a loss function; typically the mean-squared-error (MSE) between $f_\theta(\obs)$ and $\img$. 
This presents a challenge in applications where we do not have access to example images. 
Moreover, even if we have example images, we might have a large set of measurements as well, and would like to use that set to refine our reconstruction algorithm. 

In the nomenclature of machine learning, the measurements $\obs$ would be considered features and the images $\img$ the labels. Thus when $\mH=\mathbf{I}$ the training problem simplifies to learning from noisy labels. When $\mH\neq\mathbf{I}$ the training problem is to learn from noisy linear transformations of the labels.

Learning from noisy labels has been extensively studied in the context of classification; see for instance  \cite{natarajan2013learning,xiao2015learning,liu2016classification,sukhbaatar2014training,sukhbaatar2014learning}. However, the problem of learning from noisy data has been studied far less in the context of image recovery. 
In this work we show that the SURE framework can be used to 
i) denoise an image with a neural network (NN) without any training data,
ii) train NNs to denoise images from noisy training data, and 
iii) train a NN, using only noisy measurements, to solve the compressive sensing problem. 


Two concurrent and independently developed works overlap with some of our contributions.
Specifically, \cite{MCSUREtraining} demonstrates that SURE can be used to train CNN based denoisers without ground truth data and \cite{zhussip2018simultaneous} demonstrates that SURE can be used to train CNNs for compressive sensing using only noisy compressive measurements.



\section{SURE and its Generalizations} \label{sec:SURE}

The goal of this work is to reconstruct an image $\img$ from a noisy linear observations $\obs = \mH \img + \noise$, and knowledge of the linear measurement operator $\mH$. In addition to $(\obs,\mH)$, we assume that we are given training measurements $\obs_1,\obs_2,\ldots,\obs_L$ but not the images $\img_1,\img_2,\ldots,\img_L$ that produced them (we also consider the case where no training measurements are given). 
Without access to $\img_1,\img_2,\ldots,\img_L$ we cannot fit a model that minimizes the MSE, but we can minimize a loss based on Stein's Unbiased Risk Estimator (SURE).
In this section, we introduce SURE and its generalizations.

\paragraph{SURE.}  SURE is a model selection technique that was first proposed by its namesake in \cite{SURE}. SURE provides an unbiased
estimate of the MSE of an estimator of the mean of a Gaussian distributed random vector, with unknown mean. 
Let $\vx$ denote a vector we would like to estimate from noisy observations $\vy=\vx+\vw$ where $\vw\sim \mathcal{N}(0,\sigma^2 \mI)$. Also, assume $f_\theta(\cdot)$ is a weakly differentiable function parameterized by $\theta$ which receives noisy observations $\vy$ as input and provides an estimate of $\vx$ as output. Then, according to \cite{SURE}, we can write the expectation of the MSE with respect to the random variable $\noise$ as
\begin{align}\label{eqn:SURE_loss}
    \EX[\vw]{ \frac{1}{n}\|\vx-f_\theta(\vy)\|^2}
    =
    \EX[\vw]{ \frac{1}{n}\|\vy-f_\theta (\vy)\|^2}
    -\sigma_w^2
    +\frac{2\sigma_w^2}{n}\text{div}_\vy(f_\theta (\vy)),
\end{align}
where $\text{div}(\cdot)$ stands for divergence and is defined as
\begin{align}
    \text{div}_\vy(f_\theta (\vy))=\sum_{n=1}^N \frac{\partial f_{\theta n}(\vy)}{\partial y_n}.
\end{align}

Note that two terms within the SURE loss \eqref{eqn:SURE_loss} depend on the parameter $\theta$. The first term,  $\frac{1}{n}\|\vy-f_\theta (\vy)\|^2$ minimizes the difference between the estimate and the observations (bias). The second term, $\frac{2\sigma_\vw^2}{n}\text{div}_\vy(f_\theta (\vy))$ penalizes the denoiser for varying as the input is changed (variance). Thus SURE is a natural way to control the bias variance trade-off of a recovery algorithm.

The central challenge in using SURE in practice is computing the divergence. With advanced denoisers the divergence is hard or even impossible to express analytically.

\paragraph{Monte-Carlo SURE.} MC-SURE is a Monte Carlo method to estimate the divergence, and thus the SURE loss, that was proposed in \cite{MCSURE}. In particular, the authors show that for bounded functions $f_\theta$ we have
\begin{align}\label{eqn:MCdiv}
    \text{div}_\vy(f_\theta (\vy))=\lim_{\epsilon \rightarrow 0} \mathbb{E}_\vb \left\{ \vb^t \big( \frac{f_\theta(\vy+\epsilon \vb)-f_\theta(\vy)}{\epsilon} \big)\right\},
\end{align}
where $\vb$ is an i.i.d.~Gaussian distributed random vector with unit variance elements.

Following the law of large numbers, this expectation can be approximated with Monte Carlo sampling. Thanks to the high dimensionality of images, a single sample well approximates the expectation. The limit can be approximated well by using a small value for $\epsilon$; we use $\epsilon=\frac{\max(\vy)}{1000}$ throughout. This approximation leaves us with
\begin{align}\label{eqn:MCdiv_approx}
    \text{div}_\vy(f_\theta (\vy))\approx  \vb^t \left( \frac{f_\theta(\vy+\epsilon \vb)-f_\theta(\vy)}{\epsilon} \right).
\end{align}

Combining the SURE loss \eqref{eqn:SURE_loss} with the estimate of the divergence \eqref{eqn:MCdiv_approx}, enables minimization of the MSE of a denoising algorithm without ground truth data. 

\paragraph{Generalized SURE.} GSURE was proposed in \cite{GSURE} to estimate the MSE associated with estimates of $\img$ from a linear measurement $\obs = \mH \img + \noise$, where $\mH\neq \mathbf{I}$, and $\noise$ has known covariance and follows any distribution from the exponential family. 
For the special case of i.i.d.~Gaussian distributed noise the estimate simplifies to
\begin{align}\label{eqn:GSURE}
    \EX[\vw]{ \frac{1}{n}\|\mathbf{P}\img-\mathbf{P}f_\theta(\vy)\|^2}
    =\mathbb{E}_\vw {\Big[} \frac{1}{n}\|\mathbf{P}\vx\|^2+\frac{1}{n}\|\mathbf{P}f_\theta(\vy)\|^2+\frac{2\sigma_\vw^2}{n}\text{div}_\vy(f_\theta (\vy))
    -\frac{2}{n}f_\theta (\vy)^t\mH^\dagger \vy {\Big]},
\end{align}
where $\mathbf{P}$ denotes orthonormal projection onto the range space of $\mH$ and $\mH^\dagger$ is the pseudoinverse of $\mH$. Note that while this expression involves the unknown $\img$, it can be minimized with respect to $\theta$ without knowledge of $\img$.

\paragraph{Propogating Gradients.}
Minimizing SURE requires propagating gradients with respect to the Monte Carlo estimate of the divergence \eqref{eqn:MCdiv_approx}. This is challenging to do by hand, but made easy by TensorFlow's and PyTorch's auto-differentiation capabilities, which are used throughout much of our experiments below.


\section{Denoising Without Training Data\label{sec:OneShotDenoising}}

CNNs offer state-of-the-art performance for many image recovery problems including super-resolution \cite{SRCNN}, inpainting \cite{yang2017high}, and denoising \cite{DnCNN}.
Typically, these networks are trained on a large dataset of images. 
However, it was recently shown that if just fit to a single corrupted image $\vy$, without any pre-training, CNNs can still perform effective image recovery. 
Specifically,~the recent work {\it Deep Image Prior} \cite{DeepImagePrior} demonstrated that a randomly initialized CNN, trained so that its output matches the corrupted image $\vy$ well, performs exceptionally well at the aforementioned inverse problems. 
Similar to more traditional image recovery methods like BM3D \cite{BM3D}, the deep image prior method only exploits structure within a given image and does not use any external training data.


\subsection{Deep Image Prior}

The deep image prior is a CNN, denoted by $f_\theta$ and parameterized by the weight and bias vector $\theta$ that maps an input tensor $\vz$ to an image $f_\theta(\vz)$. 
The paper~\cite{DeepImagePrior} proposes to minimize
\[
E(f_\theta(\vz), \vy),
\]
over the parameters $\theta$, starting from a random initialization, using a method such as gradient descent. 
Here, $E$ is a loss function, chosen as the squared $\ell_2$-norm, i.e., $E(\vx, \vx') = \norm[2]{\vx - \vx'}^2$. 
In this paper, we consider a variant of the original Deep Image Prior work, where $\vz$ is set equal to $\vy$.\footnote{As observed in \cite{DeepImagePrior}, we found that perturbing the input slightly at every iteration helped with convergence. We will return to this observation in Section \ref{sec:deepimagepriortheory}.}

Following \cite{DeepImagePrior}, our goal is to train the network, i.e., optimize the parameters $\theta$, such that it fits the image but not the noise. 
If the network is trained using too few iterations, then the network does not represent the image well. 
If the network is trained until convergence, however, then the network's parameters are overfit and describe the noise along with the actual image.

\begin{figure}[t]
\centering
\subfigure[U-net Data Fidelity Training Loss]{\includegraphics[width=.49\textwidth]{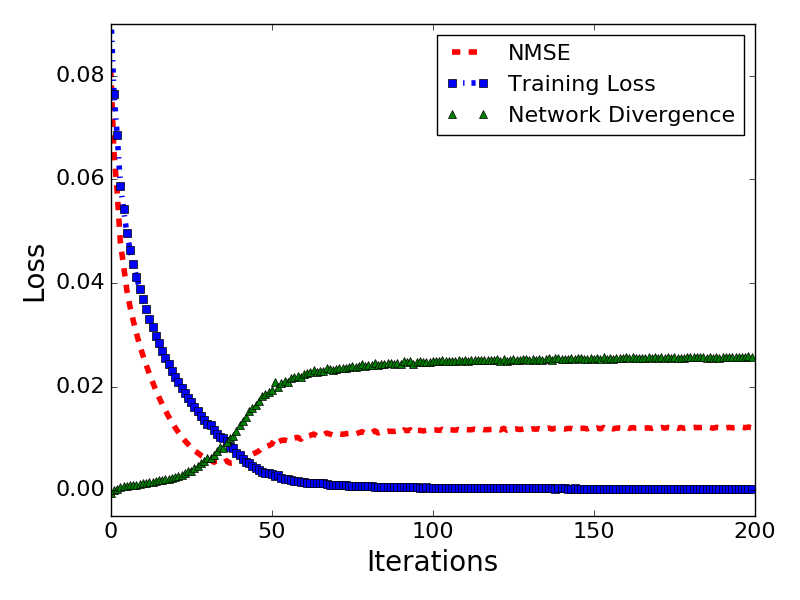}}
\subfigure[U-net SURE Training Loss]{\includegraphics[width=.49\textwidth]{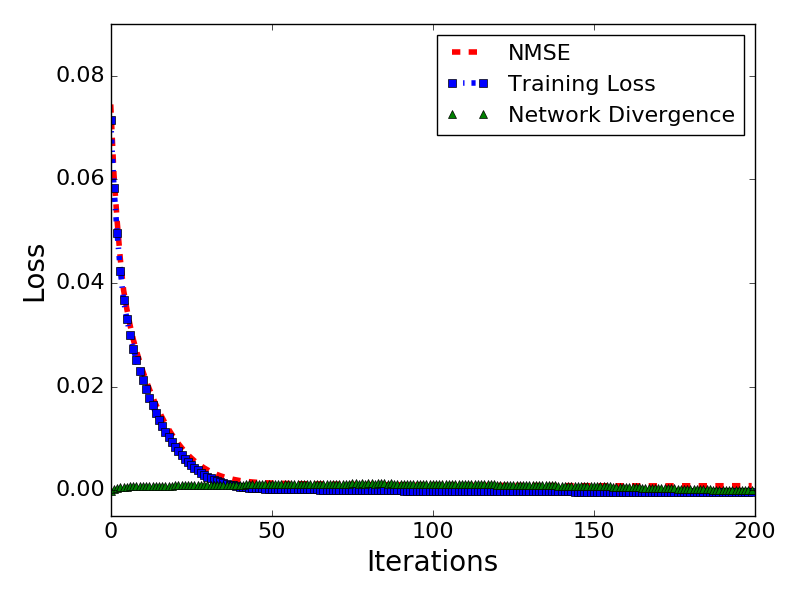}}
\caption{ The training and test errors for networks trained with $\frac{1}{n}\|\vy-f_\theta(\vy)\|^2$ Loss (a) and the SURE Loss \eqref{eqn:SURE_loss} (b). Without the SURE divergence term, the network starts to overfit and the NMSE worsens even while the training loss improves. Minimizing the SURE loss minimizes the NMSE.}
\label{fig:ReconErrs}
\end{figure}

Thus, there is a sweet spot, where the network minimizes the error between the true image and the reconstructed image.
This is illustrated in Figure~\ref{fig:ReconErrs}(a).
Figure~\ref{fig:ReconErrs}(a) displays the training loss and NMSE associated with denoising a $512\times 512$ Mandrill image which was contaminated with additive white Gaussian noise (AWGN) with standard deviation 25. It also displays the divergence of the network \eqref{eqn:MCdiv} (scaled by $2\sigma_w^2/n$). 
The figure shows that after a few iterations the network begins to overfit to the noisy image and normalized MSE (NMSE) starts to increase. 
Unfortunately, by itself the training loss gives little indication about when to stop training; it smoothly decays towards $0$.

The network, $f_\theta$, used to generate the results in this section was a very large U-net architecture \cite{Unet} with $128$ feature channels per layer. We set its input tensor $\vz$ to $\vy$; that is, the network was fed the noisy observation $\vy$ as an input. This network architecture was used in the ``flash-no flash'' experiments from \cite{DeepImagePrior}, and offers performance incrementally better than the expansive CNN fed with a random $\vz$ the authors used for denoising experiments. We used this network architecture as it allows us to compute the divergence using the Monte-Carlo approximation \eqref{eqn:MCdiv_approx}. 
Qualitatively, our network behaves the same as network originally used for denoising in \cite{DeepImagePrior}; both require early stopping to achieve optimal performance.

\subsection{SURE Deep Image Prior}

Returning to Figure \ref{fig:ReconErrs}(a), we observe that the MSE is minimized at a point where the training loss is reasonably small, but at the same time the network divergence is not too large; thus the sum of the two terms is small. 
In fact, if one plots the SURE estimate of the loss (not shown here), it lies directly on top of the observed NMSE.


Inspired by this observation, we propose to train with the SURE loss instead of the fidelity or $\ell_2$ loss. The results are shown in~Figure~\ref{fig:ReconErrs}(b). It can be seen that not only does the network avoid overfitting, the final NMSE is superior to that achieved by optimal stopping with the fidelity loss.



\begin{figure}[t]
\centering
\subfigure[Original Noisy Image]{\includegraphics[width=.24\textwidth]{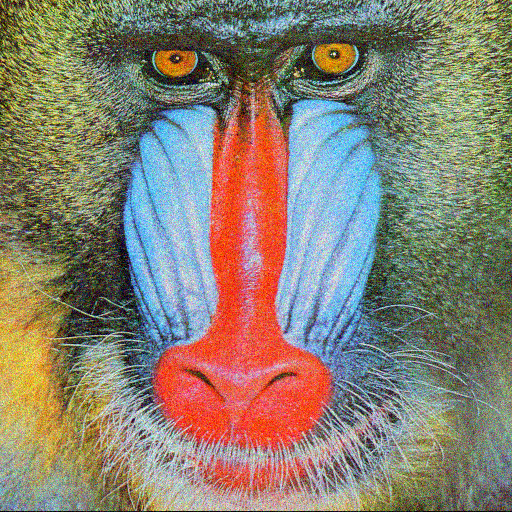}}
\subfigure[CBM3D (25.9 dB, 4.84 sec)]{\includegraphics[width=.24\textwidth]{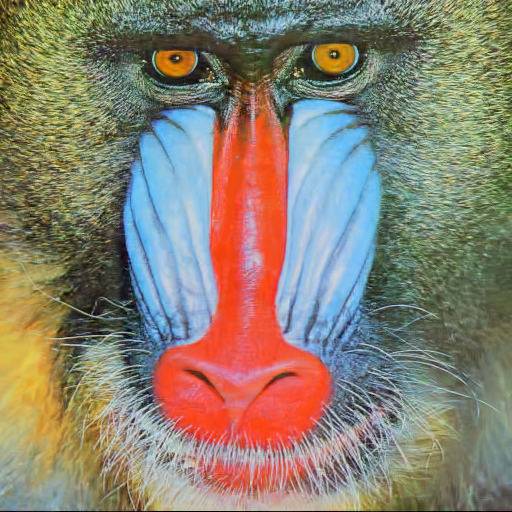}}
\subfigure[DnCNN (26.4 dB, 0.009 sec)]{\includegraphics[width=.24\textwidth]{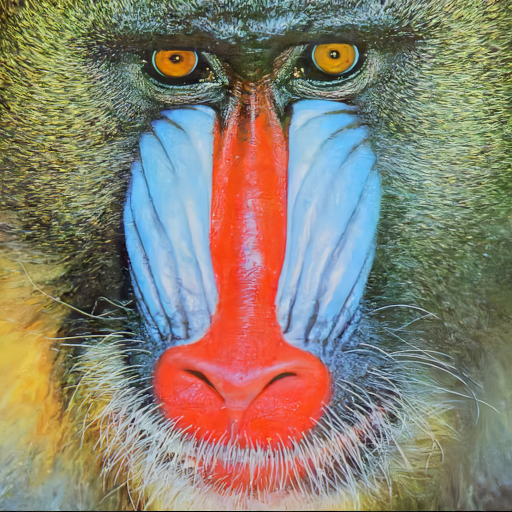}}
\subfigure[Untrained U-net (26.3 dB, 72.15 sec)]{\includegraphics[width=.24\textwidth]{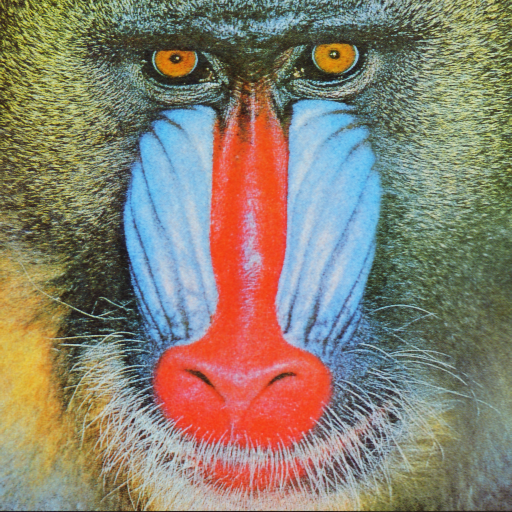}}
\caption{Reconstructions of $512 \times 512$ color Mandrill test image tested with AWGN with standard deviation 25. The untrained CNN offers performance similar to state-of-the-art methods but takes much longer.}
\label{fig:DeepPrior}
\end{figure}

We tested the performance of this large U-net ``trained'' only on the image to be denoised against state-of-the-art trained and untrained algorithms DnCNN~\cite{DnCNN} and CBM3D~\cite{BM3D}. Results are presented in Figure~\ref{fig:DeepPrior}. When trained with the SURE loss the U-net trained only on the image to be denoised offers performance competitive with these two methods. However, the figure also demonstrates that the deep image prior method is very slow and that training across additional images, as exemplified by DnCNN, is beneficial.\footnote{In this section we used the DnCNN authors' implementation of the algorithm. This implementation is faster than our own implementation, which is used elsewhere, but does not support auto-differentiation.}

The networks described in this section were trained using the Adam optimizer \cite{ADAMopt} with a learning rate of 0.001.
Throughout the paper we report recovery accuracies in terms of PSNR, defined as 
\[
\text{PSNR}=10\log_{10}\bigg(\frac{255^2}{\|\hat{x}-x\|^2}\bigg),
\]
where all images lie in the range $[0,255]$. 
All experiments were performed on a desktop with an Intel 6800K CPU and an Nvidia Pascal Titan X GPU.




\section{Denoising with Noisy Training Data}\label{sec:Denoising}

In the previous section we showed that SURE can be used to fit a CNN denoiser without training data. This is useful in regimes where no training data whatsoever is available. However, as the previous section demonstrated, these untrained neural networks are computationally very expensive to apply.

In this section we focus on the scenario where we can capture additional {\em noisy} training data with which to train a neural network for denoising. 
We show that if provided a set of $L$ \emph{noisy} images $\vy_1, \vy_2, \ldots, \vy_L$, 
training the network with the SURE loss improves upon training with the MSE loss. Specifically, we optimize the network's parameters by minimizing
\[
\sum_{\ell=1}^L  \frac{1}{n}\|\vy_\ell-f_\theta (\vy_\ell)\|^2-\sigma_w^2+\frac{2\sigma_w^2}{n}\text{div}_{\vy_\ell}\{(f_\theta (\vy_\ell))\},
\]
rather than 
\[
\sum_{\ell=1}^L \frac{1}{n}\|\vx_\ell-f_\theta(\vy_\ell)\|^2.
\]
As before we will use the Monte-Carlo estimate of the divergence \eqref{eqn:MCdiv}.

\subsection{DnCNN and Experimental Setup\label{sec:expsetupDnCNN}}

\begin{figure}[t]
\centering
\subfigure[]{\includegraphics[width=.16\textwidth]{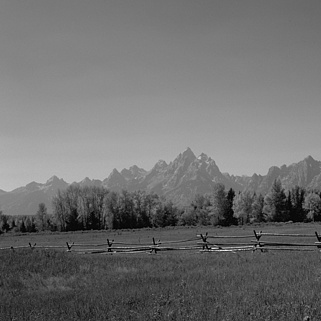}}
\subfigure[]{\includegraphics[width=.16\textwidth]{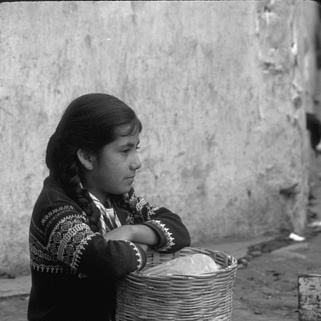}}
\subfigure[]{\includegraphics[width=.16\textwidth]{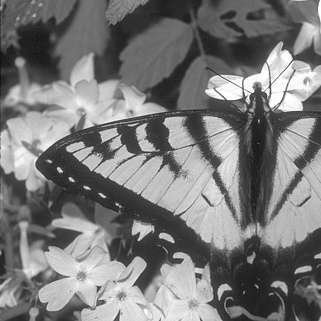}}
\subfigure[]{\includegraphics[width=.16\textwidth]{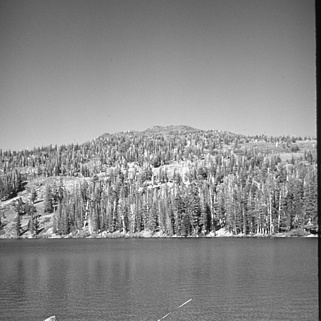}}
\subfigure[]{\includegraphics[width=.16\textwidth]{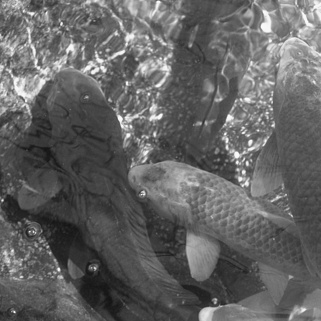}}
\subfigure[]{\includegraphics[width=.16\textwidth]{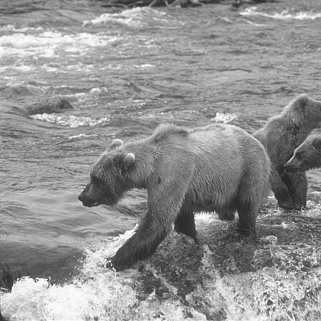}}
\caption{Six representative training images from the BSD dataset.}\label{fig:TrainingImages}

\subfigure[Barbara]{\includegraphics[width=.16\textwidth]{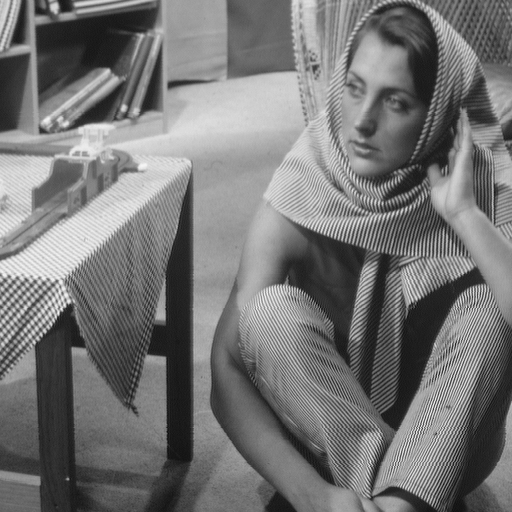}}
\subfigure[Boat]{\includegraphics[width=.16\textwidth]{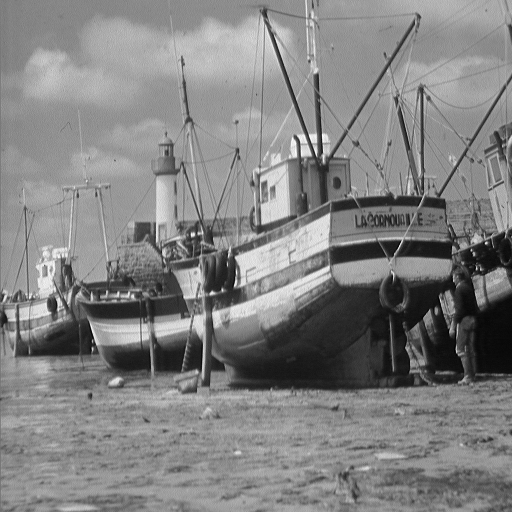}}
\subfigure[Couple]{\includegraphics[width=.16\textwidth]{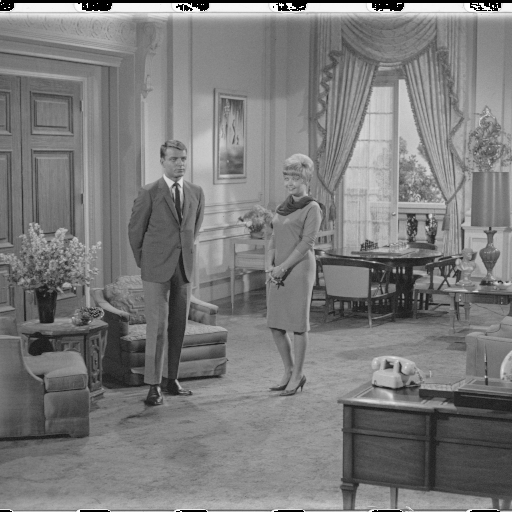}}
\subfigure[House]{\includegraphics[width=.16\textwidth]{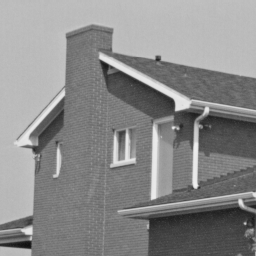}}
\subfigure[Mandrill]{\includegraphics[width=.16\textwidth]{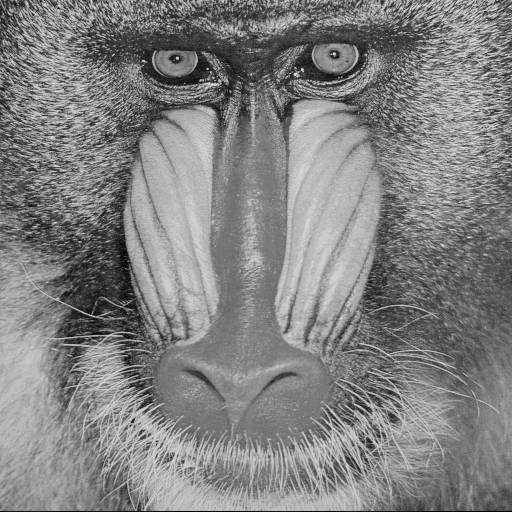}}
\subfigure[Bridge]{\includegraphics[width=.16\textwidth]{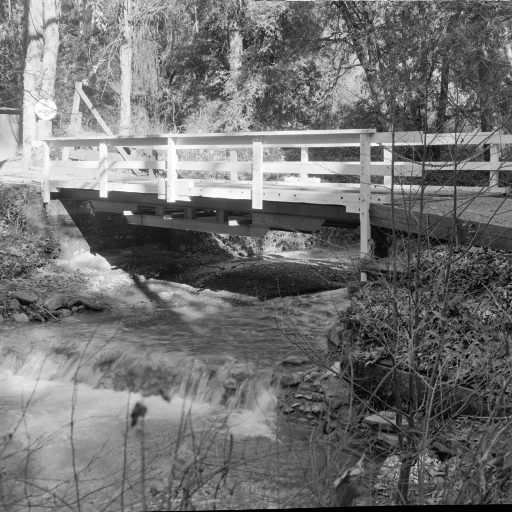}}
\caption{The six test images. They follow a similar distribution to the training images.} \label{fig:NaturalTestImages}


\end{figure}

In this section, we consider the DnCNN image denoiser, trained on grayscale images pulled from Berkeley's BSD-500 dataset~\cite{BSDDataset}, and trained using the MSE and SURE loss. Example images from this dataset are presented in Figure~\ref{fig:TrainingImages}. 
The training images were cropped, rescaled, flipped, and rotated to form a set of 204\,800 overlapping $40\times 40$ patches. 
We tested the methods on 6 standard test images presented in Figure \ref{fig:NaturalTestImages}. 

The DnCNN image denoiser \cite{DnCNN} consists of 16 sequential $3\times 3$ convolutional layers with $64$ features at each layer. Sandwiched between these layers are ReLU and batch-normalization operations. DnCNN has a single skip connection to its output and takes advantage of residual learning \cite{residuallearning}.

We trained all networks for 50 epochs using the Adam optimizer \cite{ADAMopt} with a training rate of 0.001 which was reduced to 0.0001 after 30 epochs. We used mini-batches of 128 patches.

\begin{figure}[t]
\centering
\subfigure[Original Noisy Image]{\includegraphics[width=.24\textwidth]{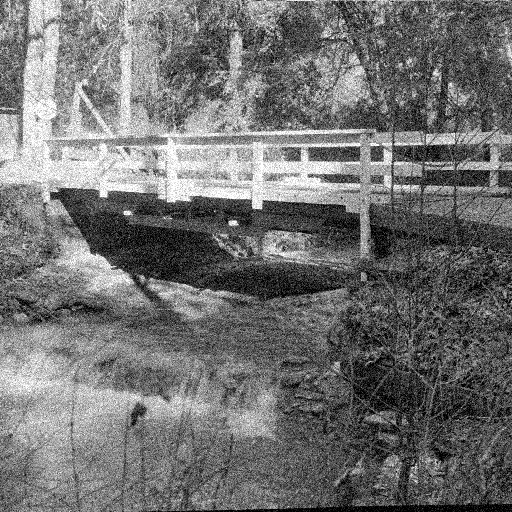}} 
\subfigure[BM3D (26.0 dB, 4.01 sec)]{\includegraphics[width=.24\textwidth]{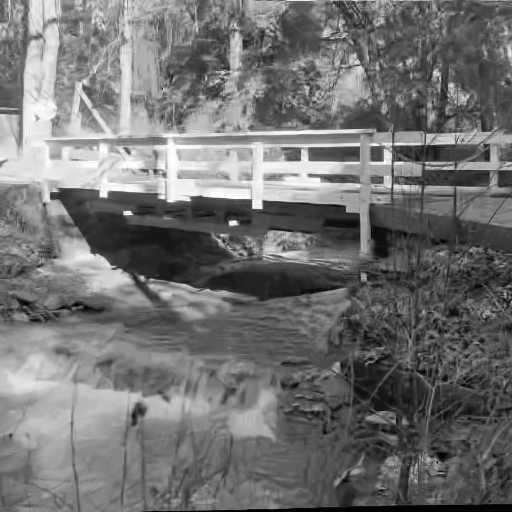}} 
\subfigure[DnCNN SURE (26.5 dB, .04 sec)]{\includegraphics[width=.24\textwidth]{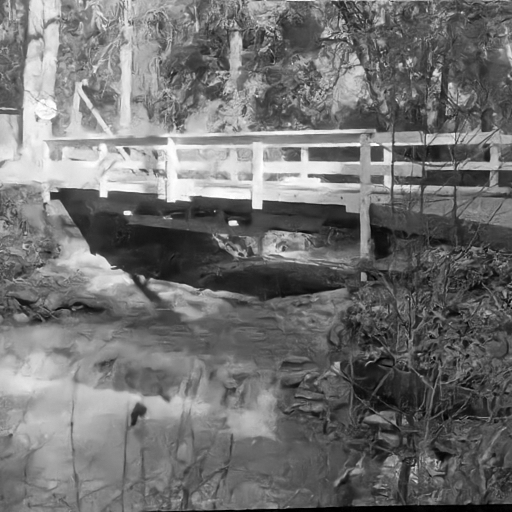}} 
\subfigure[DnCNN MSE (26.7 dB, .04 sec)]{\includegraphics[width=.24\textwidth]{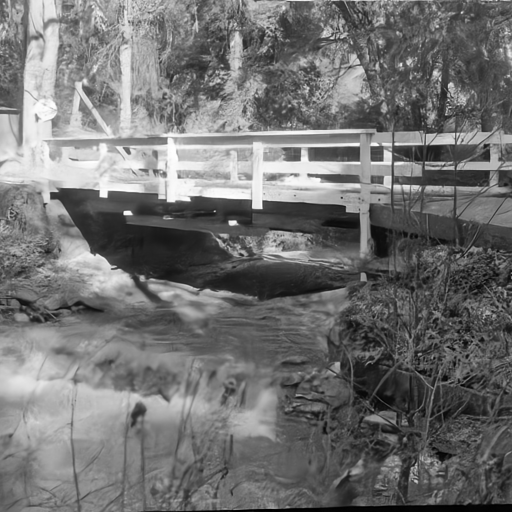}} 
\caption{Reconstructions of $512 \times 512$ grayscale Bridge test image tested with AWGN with standard deviation 25. The CNN trained with noisy data offers performance similar to state-of-the-art methods.
}
\label{fig:VisComparoDenoising}
\end{figure}

\paragraph{Results.}
The results of training DnCNN using the MSE and SURE loss are presented in Figure \ref{fig:VisComparoDenoising} and Table \ref{tab:DnCNN}. 
The results show that training DnCNN with SURE on noisy data results in reconstructions almost equivalent to training with the true MSE on clean images. Moreover, both DnCNN trained with SURE and with the true MSE outperform BM3D. As expected, because calculating the SURE loss requires two calls to the denoiser, it takes roughly twice as long to train as does training with the MSE.

\begin{table}[htbp]
  \begin{center}
    \begin{tabular}{lrrrrrrrr}
          & \multicolumn{1}{l}{\makecell{Training\\Time}} & \multicolumn{1}{l}{\makecell{Test\\Time}} & \multicolumn{1}{l}{Barbara} & \multicolumn{1}{l}{Boat} & \multicolumn{1}{l}{Couple} & \multicolumn{1}{l}{House} & \multicolumn{1}{l}{Mandrill} & \multicolumn{1}{l}{Bridge} \\
          \cmidrule{2-9}
    BM3D  & \multicolumn{1}{l}{\makecell{N/A}} & 0.82 sec & \textbf{29.8} & 28.7 & 29.2 & 32.8 & 25.8  & 26.2 \\
    \hline
    DnCNN SURE & 8.1 hrs &  0.01 sec & 29.0 & 28.9 & 29.1 & 32.3 & 26.1 & 26.6 \\
    \hline
    DnCNN MSE & 4.3 hrs & 0.01 sec & 29.4 & \textbf{29.1} & \textbf{29.5}  & \textbf{32.9 }& \textbf{26.3} & \textbf{26.7} \\
    \hline
    \end{tabular}%
    \end{center}
  \caption{\label{tab:DnCNN} Reconstruction results for $256\times 256$ images with AWGN with standard deviation 25. Even though DnCNN SURE uses only noisy training data, it performs almost as well as the DnCNN MSE training on clean data. Results were averaged over 5 trials.}
\end{table}%

\section{Compressive Sensing Recovery with Noisy Training Measurements}\label{sec:CS}
In this section we study the problem of image recovery from undersampled linear measurements. The main novelty of this section is that we do not have training labels (i.e., ground truth images). Instead and unlike conventional learning approaches in compressive sensing (CS), we train the recovery algorithm using \emph{only noisy undersampled linear measurements}.

\subsection{LDAMP}

We used the Learned Denoising-based Approximate Message Passing (LDAMP) network proposed in \cite{LDAMP} as our recovery method. LDAMP offers state-of-the-art CS recovery when dealing with measurement matrices with i.i.d.~Gaussian distributed elements. LDAMP is an unrolled version of the DAMP~\cite{DAMP} algorithm that decouples signal recovery into a series of denoising problems solved at every layer. In other words and as shown in \eqref{eqn:LDAMP}, LDAMP receives a noisy version of $\vx$ (i.e., $\vx^k+\mH^*\vz^k=\vx+\sigma \vnu$) at every layer and tries to reconstruct $\vx$ by eliminating the effective noise vector $\sigma \vnu$. 

The success of LDAMP stems from its Onsager correction term (i.e., the last term on the first line of \eqref{eqn:LDAMP}) that removes the bias from intermediate solutions. As a result, at each layer the effective noise term $\sigma\vnu$ follows a Gaussian distribution whose variance is accurately predicted by $({\sigma}^k)^2$ \cite{MalekiThesis}.

\begin{tcolorbox}
 Learned Denoising-based AMP (LDAMP) Neural Network

 For $k=1,...~K$
\begin{eqnarray}\label{eqn:LDAMP}
\vz^k&=& \vy-\mH\vx^k+\frac{1}{m}\vz^{k-1} {\rm div} D^{k-1}_{\theta^{k-1}}(\vr^{k-1}) \nonumber \\
{\sigma}^k&=& \frac{\|\vz^k\|_2}{\sqrt{m}} \nonumber \\
\vr^k&=&\vx^k+\mH^*\vz^k\nonumber\\
\vx^{k+1} &=& D^{k}_{\theta^k} (\vr^k)
\end{eqnarray}
\end{tcolorbox}

In this work, we use a $K=10$ layer LDAMP network, where each layer itself contains a $16$ layer DnCNN denoiser $D^k_{\theta^k}(\cdot)$. 
Below we use $f_\theta(\cdot)$ to denote LDAMP.

\subsection{Training LDAMP and Experimental Setup}

We compare three methods of training LDAMP. All three methods utilize layer-by-layer training, which in the context of LDAMP is minimum MSE optimal \cite{LDAMP}.

The first method, LDAMP MSE, simply minimizes the MSE with respect to the training data.
The second method, LDAMP SURE, uses SURE to train LDAMP using only noisy measurements. 
This method takes advantage of the fact that at each layer LDAMP is solving denoising problems with known variance ${{\sigma}^k}^2$.
\[
\theta_{\text{SURE}}=\arg\min_\theta \sum_{\ell=1}^L  \frac{1}{n}\|\vr^K_\ell-f_\theta (\vy_\ell)\|^2-{{\sigma}^K}^2+\frac{2{({\sigma}^K})^2}{n}\text{div}_{\vy_\ell}(f_\theta (\vy_\ell)).
\]

The third and final method, LDAMP GSURE, uses generalized SURE to train LDAMP using only noisy measurements.  
\[
\theta_{\text{GSURE}}=\arg\min_\theta \sum_{\ell=1}^L  \frac{1}{n}\|\mathbf{P}f_\theta(\vy_\ell)\|^2+\frac{2\sigma_\vw^2}{n}\text{div}_\vy(f_\theta (\vy_\ell))
-\frac{2}{n}f_\theta (\vy_\ell)^t\mH^\dagger \vy_\ell.
\]

The SURE and GSURE methods both rely upon Monte-Carlo estimation of the divergence \eqref{eqn:MCdiv}.

We used $m\times n$ dense Gaussian measurement matrices for our low resolution training and modified ``fast JL transform'' matrices \cite{ailon2006approximate}, which offer $O(n\log n)$ multiplications, for our high resolution testing.\footnote{Our approximate ``fast JL transform'' matrices were of the form $\mathbf{H}=\mathbf{P}\mathbf{F}\mathbf{D}$ where $\mathbf{P}$ is a sparse $m\times n$ matrix formed by randomly sampling the rows of an $n\times n$ identity matrix, $\mathbf{F}$ is a Discrete Cosine Transform matrix, and $\mathbf{D}$ is a diagonal matrices with i.i.d.~entries that take values $-\sqrt{\frac{n}{m}}$ and $\sqrt{\frac{n}{m}}$ with probability $.5$.} For both training and testing we sampled with $\frac{m}{n}=.2$ and $\sigma_w=1$. See \cite{LDAMP} for more details about the measurement process. Other experimental settings such as batch sizes, learning rates, etc. are the same as those in Section~\ref{sec:expsetupDnCNN}.

\begin{figure}[t]
\centering
\subfigure[BM3D-AMP (30.2 dB, 16.5 sec)]{\includegraphics[width=.24\textwidth]{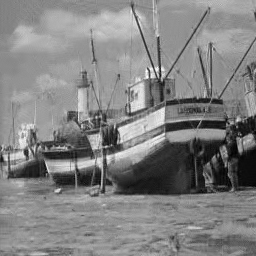}} 
\subfigure[LDAMP GSURE (24.7 dB, 0.4 sec)]{\includegraphics[width=.24\textwidth]{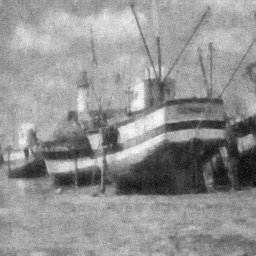}} 
\subfigure[LDAMP SURE (29.5 dB, 0.4 sec)]{\includegraphics[width=.24\textwidth]{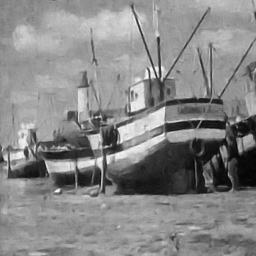}} 
\subfigure[LDAMP MSE (31.0 dB, 0.4 sec)]{\includegraphics[width=.24\textwidth]{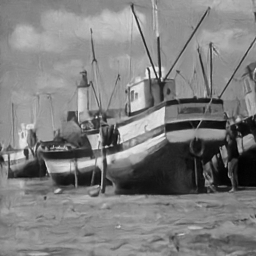}} 
\caption{Reconstructions of $256 \times 256$ Boat test image sampled at a rate of $\frac{m}{n}=0.2$ using modified fast JL transform matrices and i.i.d. Gaussian distributed measurement noise with standard deviation 1. LDAMP trained with SURE offers performance on par with BM3D-AMP. 
}
\label{fig:VisualComparoCS}
\end{figure}

\paragraph{Results.}
The networks resulting from training LDAMP with the MSE, SURE, and GSURE losses are compared to BM3D-AMP \cite{DAMP} in Figure \ref{fig:VisualComparoCS} and Table \ref{tab:CS1}. The results demonstrate LDAMP networks trained with SURE and MSE both perform roughly on par with BM3D-AMP and run significantly faster. 
The results also demonstrate that LDAMP trained with GSURE offers significantly reduced performance. This can be understood by returning to the original GSURE cost \eqref{eqn:GSURE}. Minimizes GSURE minimizes the distance between $\mathbf{P}\vx$ and $\mathbf{P}f_\theta(\vy)$ not between $\vx$ and $f_\theta(\vy)$, where $\mathbf{P}$ denotes orthogonal projection onto the range space of $\mathbf{H}$.
In the context of compressive sensing, the range space is small and so these two distances are not proportional to one another.


\begin{table}[t]
  \begin{center}
    \begin{tabular}{crrrrrrrr}
          & \multicolumn{1}{l}{\makecell{Training\\Time}} & \multicolumn{1}{l}{\makecell{Test\\Time}} & \multicolumn{1}{l}{Barbara} & \multicolumn{1}{l}{Boat} & \multicolumn{1}{l}{Couple} & \multicolumn{1}{l}{House} & \multicolumn{1}{l}{Mandrill} & \multicolumn{1}{l}{Bridge} \\
          \cmidrule{2-9}
    \makecell{BM3D-AMP}  & N/A   & 16.5 sec & \textbf{31.3 } & 30.3& \textbf{31.6} & \textbf{38.1}& 24.7 & 25.5 \\
    \hline
    \makecell{LDAMP\\SURE} & 34.7 hrs & 0.4 sec & 29.1 & 29.5 & 29.5 & 35.2 & 24.4 & 25.5\\
    \hline
    \makecell{LDAMP\\GSURE} & 43.2 hrs & 0.4 sec & 25.2 & 24.5 & 24.7 & 28.2 & 23.0 & 22.9 \\
    \hline
    \makecell{LDAMP\\MSE} & 20.7 hrs & 0.4 sec & 30.2 & \textbf{31.0} & \textbf{31.6} &  36.4 & \textbf{24.9} & \textbf{25.9}\\
    \hline
    \end{tabular}%
  \end{center}
  
  \caption{\label{tab:CS1}Reconstruction results for $256\times 256$ images sampled at $\frac{m}{n}=0.2$, using modified fast JL transform matrices with i.i.d.~Gaussian noise with variance $1$. Results were averaged over 5 trials.}
  
\end{table}%

\section{Related Work}\label{sec:relatedowork}
Denoising and compressive sensing are each mature problems with vast literature. We list some of the most relevant works here.

\paragraph{Denoising with SURE}
SURE has a a long history in the context of image denoising. For instance, it is the key ingredient in the well-known SureShrink wavelet denoising method \cite{SureShrink}, which uses SURE to set parameters in a wavelet-domain thresholding algorithm. SureShrink preceded a number of other wavelet-thresholding denoisers that have improved upon this basic idea \cite{SURELet,raphan2008optimal}. 
SURE and related techniques have has also been generalized and applied to Poisson \cite{PURELet}, multiplicative \cite{panisetti2014unbiased}, and other distributions \cite{gubbi2014risk} of noise.

\paragraph{Denoising with Noisy Training Data}
In \cite{moon2016neural}, the authors used unbiased estimates of risk to train a neural network to denoise discrete-valued signals. In subsequent work, the authors used SURE to train a very simple neural network to denoise images \cite{cha2018neural}. 
\cite{bora2018ambientgan} used noisy, blurry, and occluded images to train a GAN that could produce noise-free images.   
\cite{Noi2noi} demonstrated CNNs can be trained to restore images using {\em pairs} of corrupted training data, even without knowing the distribution of the noise. 







\paragraph{Compressive Sensing}
The compressive sensing training/reconstruction method proposed in Section \ref{sec:CS} is closely related to the projected GSURE method \cite{giryes2011projected} and Parameterless Optimal AMP \cite{ParameterlessAMP,bayati2013estimating},  which use the (G)SURE loss to tune parallel coordinate descent and AMP \cite{AMP} algorithms, respectively, to reconstruct signals with unknown sparsity in a known dictionary. The method is also closely related to Parametric SURE AMP \cite{DaviesSUREAMP} which uses SURE to adapt a parameterized denoiser at every iteration of the AMP algorithm.

The proposed method is somewhat related to blind CS \cite{BlindCS} wherein signals that are sparse with respect to an unknown dictionary are reconstructed from compressive measurements. 
At a high level, the proposed method can be considered a practical form of universal CS \cite{jalali2014minimum,BaronUniversalCS1, BaronUniversalCS2,jalali2017universal} wherein a signal's distribution is estimated from noisy linear measurements and then used to reconstruct the signal. 

\paragraph{Other Applications}
We note in passing that unbiased risk estimators have seen use in a number of other applications including least squares estimation \cite{raphan2011least}, kernel regression \cite{krishnan2014spatially} and channel estimation \cite{upadhya2016risk}.

\section{Interpreting the Deep Image Prior}\label{sec:deepimagepriortheory}

The original Deep Image Prior paper set out to minimize
\begin{align}\label{eqn:DeepImagePriorLoss}
    \|\vy-f_\theta(\vz)\|^2,
\end{align}
and attributed the success of the proposed method at least partly to the fact that the network fit the noise slower than than it fit the signal~\cite{DeepImagePrior}.

To minimize the cost function, the paper used an iterative first order method, specifically the ADAM optimizer. However, if tuned appropriately, one can use use gradient descent and get essentially equivalent performance.

In either case, noise regularization dramatically improves the performance of the network. That is, \cite{DeepImagePrior} found it beneficial to add a noise vector $\gamma$, with $\gamma\sim N(0,\sigma_\gamma^2\mathbf{I})$, at every iteration of the optimization.
With this noise term, the optimization becomes equivalent to solving the stochastic optimization problem
\begin{align}\label{eqn:DeepImagePriorLoss_stochastic}
    \arg\min_\theta \mathbb{E}_\gamma\|\vy-f_\theta(\vz+{\gamma})\|^2.
\end{align}
This has important implications. Using the techniques that \cite{alain2014regularized} used to analyze denoising auto-encoders, we can take a Taylor expansion of $f_\theta$ around $\vz$ to express the loss function in \eqref{eqn:DeepImagePriorLoss_stochastic} as
\begin{align*}
    \mathbb{E}_\gamma&\Big\|\vy-f_\theta(\vz)+\frac{\partial f_\theta(\vz)}{\partial \vz}{\gamma}+o(\sigma_\gamma^2)\Big\|^2&\nonumber\\
    =&\mathbb{E}_\gamma\Big[\|\vy-f_\theta(\vz)\|^2-2(\vz-f_\theta(\vz))^t\frac{\partial f_\theta(\vz)}{\partial \vz}{\gamma}-2(\vz-f_\theta(\vz))^to(\sigma_\gamma^2)\nonumber\\
    &+\gamma^t\frac{\partial f_\theta(\vz)}{\partial z}^t\frac{\partial f_\theta(\vz)}{\partial \vz}\gamma+2\gamma^t\frac{\partial f_\theta(\vz)}{\partial z}^to(\sigma_\gamma^2)+o(\sigma_\gamma^4)\Big]\\
    =&\|\vy-f_\theta(\vz)\|^2-2(\vz-f_\theta(\vz))^t\frac{\partial f_\theta(\vz)}{\partial \vz}\mathbb{E}_\gamma[{\gamma}]
    +{\rm Tr}\Big(\mathbb{E}_\gamma[\gamma\gamma^t]\big[\frac{\partial f_\theta(\vz)}{\partial z}^t\frac{\partial f_\theta(\vz)}{\partial \vz}\big]\Big)+o(\sigma_\gamma^2).
\end{align*}

Which, by substituting in $\mathbb{E}_\gamma[\gamma]=\mathbf{0}$ and $\mathbb{E}_\gamma[\gamma\gamma^t]=\sigma_\gamma^2\mathbf{I}$, simplifies to
\begin{align}\label{eqn:DAEloss}
    \|\vy-f_\theta(\vz)\|^2+\sigma^2\Big\|\frac{\partial f_\theta(\vz)}{\partial \vz}\Big\|^2_F+o(\sigma_\gamma^2) \text{ as } \sigma_\gamma\rightarrow 0.
\end{align}

As a consequence, the deep image prior can be interpreted as minimizing 
the loss function \eqref{eqn:DAEloss}, as opposed to the loss function \eqref{eqn:DeepImagePriorLoss}. 

When $\vz=\vy$, as was the case in Section \ref{sec:OneShotDenoising}, this result can be understood geometrically. The Frobenius norm of the Jacobian penalizes the curvature of the loss function around $\vy$: it encourages the vector field $f_\theta$ to be small around $\vy$.

In contrast, training a network with the SURE loss, which does not require jittering the input, minimizes the cost
\begin{align}\label{eqn:SUREsingleloss}
    \|\vy-f_\theta(\vy)\|^2+2\sigma_w^2\sum_{n=1}^N \frac{\partial f_{\theta n}(\vy)}{\partial y_n}.
\end{align}
SURE penalizes the trace of the Jacobian. Geometrically, such a penalty encourages $f_\theta$ to be a sink at $\vy$.


In either case, given enough parameters the network should be able to simultaneously set the data fidelity term $\|\vy-f_\theta(\vy)\|^2$ to $0$, that is fit the noisy signal $\vy$ exactly, while minimizing the regularization term. In fact, with the SURE loss, the network could ignore the data fidelity term entirely and focus on driving the divergence to negative infinity. Fortunately, in practice the learning dynamics are such that the solutions recovered by the network are stable for many hundreds of iterations.







\section{Conclusions}

We have made three distinct contributions. 
First we showed that SURE can be used to denoise an image using a CNN without any training data. 
Second, we demonstrated that SURE can be used to train a CNN denoiser using only noisy training data. 
Third, we showed that SURE can be used to train a neural network, using only noisy measurements, to solve the compressive sensing problem. 

In the context of imaging, our work suggests a new hands-off approach to reconstruct images. Using SURE, one could toss a sensor into a novel imaging environment and have the sensor itself figure out and then apply the appropriate prior to reconstruct images.

In the context of machine learning, our work suggests that divergence may be an overlooked proxy for variance in an estimator. Thus, while SURE is applicable in only fairly specific circumstances, penalizing divergence could be applied more broadly as a tool to help attack overfitting.




{\small
	\bibliographystyle{./IEEEtran}
	\bibliography{./refs}
}







\end{document}